\documentclass{article}

\usepackage[preprint]{neurips_2026}

\usepackage[utf8]{inputenc} 
\usepackage[T1]{fontenc}    
\usepackage{hyperref}       
\usepackage{url}            
\usepackage{booktabs}       
\usepackage{amsfonts}       
\usepackage{nicefrac}       
\usepackage{microtype}      
\usepackage{xcolor}         

\usepackage{amsmath}
\usepackage{txfonts}
\usepackage{url}
\usepackage{algpseudocode}
\usepackage{tikz}
\usepackage{float}
\usepackage{multirow}
\usepackage{csquotes}
\usepackage{subcaption}
\usetikzlibrary{positioning, arrows}

\usepackage{wrapfig}

\usepackage[normalem]{ulem}
\useunder{\uline}{\ul}{}

\usepackage{graphicx}
\usepackage[noabbrev]{cleveref}

\newcommand\marcin[1]{{\color{blue} #1}}

\def\our{SoftSAE}

\title{\our{}: Dynamic Top-K Selection for Adaptive Sparse Autoencoders}

\author{%
  Jakub St\k{e}pie\'n \\
  Jagiellonian University \\
  \texttt{jaku8.stepien@student.uj.edu.pl} \\
  \and
  Marcin Mazur \\
  Jagiellonian University \\
  \texttt{marcin.mazur@uj.edu.pl} \\
  \and
  Jacek Tabor \\
  Jagiellonian University \\
  \texttt{jacek.tabor@uj.edu.pl} \\
  \and
  Przemys{\l}aw Spurek \\
  Jagiellonian University \\
  IDEAS Research Institute\\
  \texttt{przmeyslaw.spurek@uj.edu.pl} \\
}

\begin{document}

\maketitle

\vspace{-0.5cm}
 \begin{figure}[h!]
     \centering
       \includegraphics[width=1\textwidth]{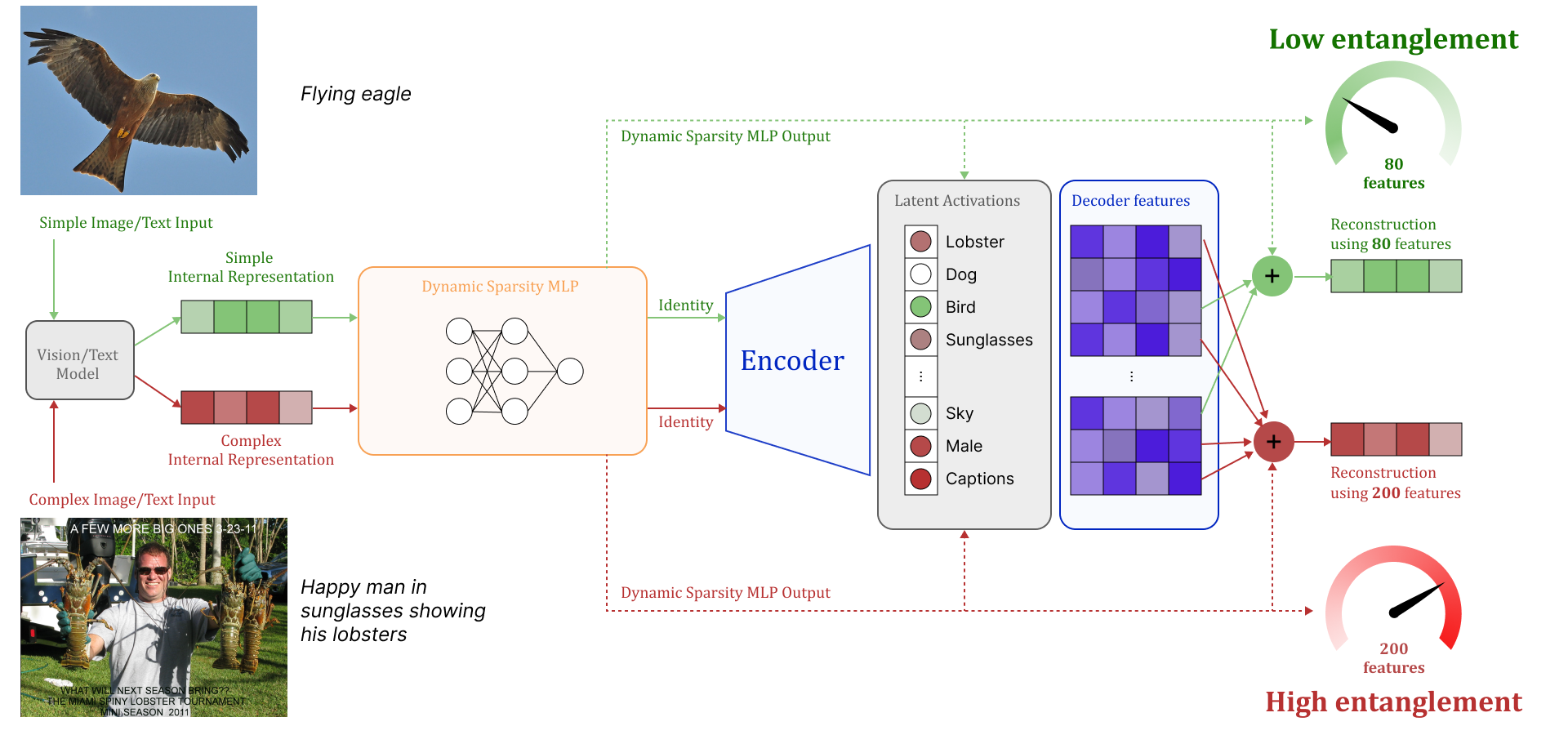}
  \vskip2mm
     \caption{
     Overview of the \our{} framework for adaptive mechanistic interpretability. Unlike traditional SAEs that enforce a fixed sparsity level \(K\) regardless of input complexity, \our{} uses a Dynamic Sparsity MLP to estimate an input-dependent sparsity \(\hat{k}\). Simple inputs, such as a ``flying eagle'', are represented with fewer active features, reducing noise and entanglement. In contrast, semantically rich or cluttered inputs, such as ``a happy man in sunglasses showing his lobsters'', induce higher sparsity budgets, enabling the model to capture multiple concepts without information loss. This adaptive mechanism aligns the effective explanation length with the local intrinsic dimensionality of the data manifold.}
  \label{teaser} 
 \label{fig:teaser} 
 \end{figure}

\begin{abstract}

Sparse Autoencoders (SAEs) have become an important tool in mechanistic interpretability, helping to analyze internal representations in both Large Language Models (LLMs) and Vision Transformers (ViTs). By decomposing polysemantic activations into sparse sets of monosemantic features, SAEs aim to translate neural network computations into human-understandable concepts. However, common architectures such as TopK SAEs rely on a fixed sparsity level. They enforce the same number of active features ($K$) across all inputs, ignoring the varying complexity of real-world data.
Natural data often lies on manifolds with varying local intrinsic dimensionality, meaning the number of relevant factors can change significantly across samples. This suggests that a fixed sparsity level is not optimal. Simple inputs may require only a few features, while more complex ones need more expressive representations. Using a constant $K$ can therefore introduce noise in simple cases or miss important structure in more complex ones.
To address this issue, we propose \our{}, a sparse autoencoder with a Dynamic Top-K selection mechanism. Our method uses a differentiable Soft Top-K operator to learn an input-dependent sparsity level $\hat{k}$. This allows the model to adjust the number of active features based on the complexity of each input. As a result, the representation better matches the structure of the data, and the explanation length reflects the amount of information in the input. Experimental results confirm that \our{} not only finds meaningful features, but also selects the right number of features for each concept. The source code is available at: \url{https://github.com/St0pien/SoftSAE}.

\end{abstract}


\section{Introduction}

Mechanistic interpretability seeks to reverse engineer the opaque representations of deep neural networks into human-understandable components. As both Large Language Models and Vision Transformers scale in complexity, understanding their internal reasoning becomes critical for safety and alignment. In recent years, Sparse Autoencoders (SAEs) \cite{huben2023sparse,zhang2024beyond} have emerged as the dominant paradigm for this task. By mapping dense, polysemantic activations into a higher-dimensional space under a strict sparsity constraint, SAEs successfully isolate monosemantic features. Early foundational approaches, such as the ReLU SAE \cite{bricken2023monosemanticity}, have demonstrated that neural networks naturally learn superposition, in which multiple distinct concepts share the same latent neurons. 

To refine feature extraction and address optimization instabilities, the research community has proposed several structural innovations. The TopK SAE~\cite{gao2024scaling} introduced a hard thresholding mechanism that explicitly activates only the most prominent features, effectively bypassing the tuning issues of continuous penalty terms. This concept was further advanced by BatchTopK \cite{bussmann2024batchtopk} to ensure better feature utilization and prevent dead neurons across diverse training batches. Concurrent innovations include the JumpReLU SAE \cite{rajamanoharan2024jumping}, which employs a discontinuous activation function to cleanly separate true signals from noise, and Matryoshka representations \cite{clip_matryoshka}, which organize learned features into highly scalable hierarchical structures. Despite impressive advancements, all these leading methodologies share a fundamental architectural limitation: they impose a rigid, global capacity constraint across the entire dataset.

Natural data exhibits immense variability in complexity. As demonstrated by Local Intrinsic Dimensionality (LID) estimation techniques such as LIDL \cite{tempczyk2022lidl}, the number of latent degrees of freedom fluctuates significantly across different regions of the data manifold. A simple linguistic conjunction or a visually uniform background requires very few active concepts to be fully described. Conversely, highly specialized domain terminology or dense, cluttered visual scenes demand a rich, multifaceted set of explanatory features. Imposing a uniform global parameter \(K\), representing the number of active features, explicitly disregards this fundamental geometric property of the data. It forces the interpretability model into an unavoidable compromise: it either generates overly complex explanations by hallucinating noisy features for simple inputs, or produces insufficient explanations by dropping critical semantic nuance for complex thoughts.

To overcome this rigid bottleneck, we propose \our{} (see \Cref{teaser}), a highly adaptive architecture that dynamically calibrates its capacity to match the exact complexity of each processed input. Our core contribution is the integration of a Dynamic Top-K Selection mechanism powered by the differentiable Soft Top-K operator \cite{struski2025lapsum}. Instead of relying on a static hyperparameter, \our{} optimizes an input-dependent scalar \(\hat{k}\), representing the sparsity level of the processed data, through continuous gradient descent. This method allows the network to learn exactly how many concepts are required to faithfully reconstruct a given representation, thereby aligning the explanation length with the signal's true information density.

The primary contributions of our work are as follows:
\vspace{-0.3cm}
\begin{itemize}
    \item We introduce \our{}, a novel approach that uses a continuous Soft Top-K formulation to dynamically predict and optimize the exact sparsity level (\(\hat{k}\)) for each input.
    \vspace{-0.2cm}
    \item We provide a strong theoretical grounding for adaptive sparsity in mechanistic interpretability by linking explanation capacity directly to the local intrinsic dimensionality of the neural activation space.
    \vspace{-0.2cm}
    \item By dynamically adjusting \(\hat{k}\), the model achieves a superior balance between reconstruction fidelity and interpretability, uncovering monosemantic features that more faithfully reflect the multiscale complexity of natural data.
\end{itemize}

\section{Related Work}

As deep learning architectures achieve unprecedented performance across diverse domains, their inherent opacity remains a significant barrier to safe and reliable deployment. The pursuit of illuminating these opaque systems has fostered a rich landscape of research, generally bifurcating into structural and semantic methodologies. In this section, we review the foundational and contemporary literature that contextualizes our proposed dynamic framework. We begin by examining the rapid evolution of mechanistic interpretability and sparse autoencoders, which strive to decode internal network representations at a highly granular level. Following this, we explore concept-based explainability paradigms that map latent geometric spaces to human-comprehensible concepts. Finally, we review recent developments in adaptive capacity and differentiable selection, positioning our methodology as a necessary synthesis of these distinct research trajectories.

\paragraph{Mechanistic Interpretability and Sparse Autoencoders}
The field of mechanistic interpretability aims to reverse engineer the opaque inner workings of complex neural networks into understandable, logical components \cite{conmy2023towards, bereska2024mechanistic}. Initial attempts in this domain focused on generating natural-language descriptions of individual neurons \cite{hernandez2021natural}. However, the highly polysemantic nature of these representations, where a single neuron responds to multiple unrelated concepts, posed significant challenges.

To disentangle these overlapping representations, Sparse Autoencoders (SAEs) emerged as a robust solution capable of recovering monosemantic features \cite{huben2023sparse, zhang2024beyond}. The foundational ReLU-based architectures \cite{bricken2023monosemanticity} paved the way for scaling these methods to massive language models \cite{templeton2024scaling}. Subsequent architectural improvements focused heavily on refining the sparsity constraint to improve both reconstruction fidelity and feature purity. The applied Top-K framework \cite{gao2024scaling} introduced a strict activation limit to bypass the instability of continuous penalty terms. This was later optimized across training batches via BatchTopK \cite{bussmann2024batchtopk} to prevent dead neurons. Other notable variants include the JumpReLU architecture \cite{rajamanoharan2024jumping}, which implements a discontinuous activation function to cleanly separate signal from noise, and Matryoshka representations \cite{clip_matryoshka}, which explore scalable hierarchical feature structuring. These advanced techniques have been successfully applied across a wide range of domains, including diffusion models \cite{surkov2024one}, radiology report generation \cite{abdulaal2024x}, and broad vision architectures \cite{bhalla2024interpreting, cywinski2025saeuron}.

\paragraph{Concept Based Explainability}
Parallel to mechanistic approaches, concept-based explainability seeks to map latent spaces to predefined, human-coherent ideas. Early frameworks relied heavily on manually annotated datasets to construct interpretable basis decompositions and concept activation vectors \cite{kim2018interpretability, zhou2018interpretable}. To scale these methods beyond small datasets, researchers developed automated concept extraction techniques \cite{ghorbani2019towards, bykov2023labeling}. Further structural innovations led to the introduction of Concept Bottleneck Models \cite{koh2020concept}, which explicitly enforce conceptual representations during training. Recent advancements have addressed the traditional accuracy-versus-interpretability trade-off by embedding continuous concept representations directly into the network \cite{espinosa2022concept}.

\begin{figure}[!h]
    \centering
    \includegraphics[width=1\linewidth]{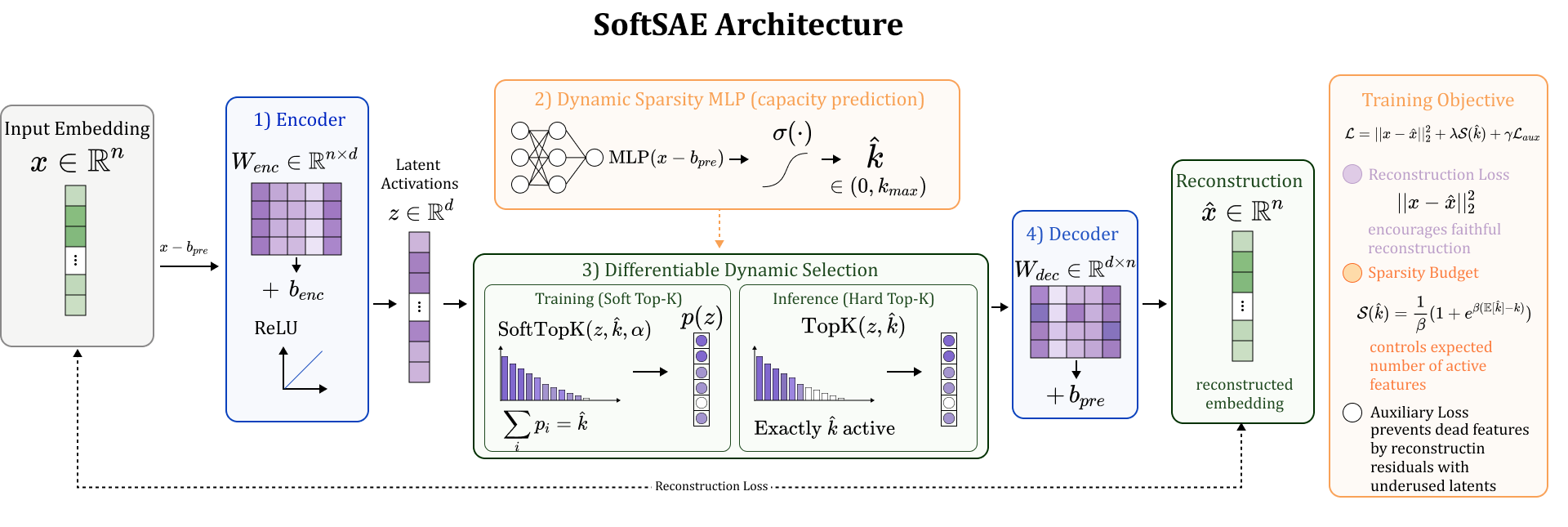}
    \caption{Diagrammatic representation of the SoftSAE architecture. The model utilizes a Dynamic Sparsity MLP to estimate an input-dependent sparsity level \(\hat{k}\). During training, a differentiable Soft Top-K operator enables gradient flow through the selection mechanism, while a hard Top-K is used at inference to ensure exact sparsity and interpretability.}
    \label{fig:architecture}
\end{figure}

\paragraph{Dynamic Capacity and Differentiable Selection}
While existing sparse architectures and concept-based models provide valuable insights, they typically enforce a static capacity parameter across all inputs, disregarding the varying complexity of individual data samples. Our approach diverges from these static constraints by drawing inspiration from differentiable selection operators. By leveraging the continuous Soft Top-K formulation \cite{struski2025lapsum}, we introduce an adaptive sparsity mechanism. Unlike rigid models, this dynamic selection allows the network to autonomously determine the optimal number of active features for each input, bridging the gap between fixed-capacity autoencoders and the inherent multiscale complexity of natural data.





\section{Our Method}

In this section, we detail the theoretical foundation and practical implementation of our \our{} architecture. We begin by formalizing the standard interpretability paradigm and identifying the mathematical limitations of static capacity constraints. Following this, we introduce the core components of our proposed model, highlighting the dedicated neural module that estimates input-dependent capacity. We then explain the integration of a differentiable Soft Top-K operator, which enables continuous, gradient-based optimization during training while ensuring strict interpretability through exact selection at inference time. Finally, we formulate the comprehensive training objective and outline a novel sparsity budget mechanism that balances reconstruction fidelity with dynamic feature allocation.
The overall \our{} architecture is visually summarized in \Cref{fig:architecture}.

\paragraph{Preliminaries and Background}
Standard interpretability techniques analyze neural model activations \(x \in \mathbb{R}^n\) extracted from a specific layer of a pretrained network. Sparse autoencoders decompose these activations into sparse linear combinations of learned directions, aiming to achieve high interpretability and monosemanticity. 

The foundational SAE's architecture consists of an encoder, a decoder, and an expanded latent space $\mathcal{Z}=\mathbb{R}^d$, where the dimension \(z\) is typically much larger than \(n\). Formally, given a pre-encoder bias \(b_{pre} \in \mathbb{R}^n\), an encoder weight matrix \(W_{enc} \in \mathbb{R}^{n \times d}\), and an encoder bias \(b_{enc} \in \mathbb{R}^d\), the latent representation \(z \in \mathbb{R}^d\) is traditionally computed using a non linear activation such as ReLU:
\begin{equation}
    z = \text{ReLU}(W_{enc}(x - b_{pre}) + b_{enc}).
\end{equation}
The input is then reconstructed using a decoder matrix \(W_{dec} \in \mathbb{R}^{d \times n}\) as follows:
\begin{equation}
    \hat{x} = W_{dec}z + b_{pre}.
\end{equation}
The rows of \(W_{dec}\) form a dictionary of \(d\) feature vectors in \(\mathbb{R}^n\), and the latent vector \(z\) specifies a linear commbination of these features used to reconstruct \(x\), accordingly \(d\) is often referred to as the dictionary size.

The basic objective during training is to minimize the reconstruction loss, defined as the squared Euclidean distance between the original and reconstructed activations: 
\begin{equation}
    \mathcal{L}(x) := ||x - \hat{x}||^2_2,
\end{equation}
often accompanied by a sparsity penalty.

While foundational models utilize continuous activations to induce sparsity, recent advancements have replaced ReLU with a rigid Top-K selection to guarantee an exact number of active concepts. However, locking this capacity to a single global value completely ignores the varying conceptual density of individual inputs, forcing the network into a suboptimal regime in which it cannot autonomously adjust its explanatory bandwidth.

The above-mentioned issue is partially mitigated by BatchTopK and Matryoshka sparse autoencoders, in which sparsity is enforced through a global threshold rather than a fixed per-input budget. During training, this threshold is estimated using an exponential moving average (EMA) of activation statistics, and then applied at inference time. As a result, the number of active features per input can vary, but this variation is largely incidental rather than a principled adaptation to input complexity. In particular, multiple high activations do not necessarily indicate that all corresponding concepts should be used, as they may reflect redundancy or noise rather than genuinely distinct explanatory factors.

\paragraph{\our{} Architecture Overview}

To address the limitations of static capacity, we propose the \our{} architecture. Our model enhances the foundational autoencoder structure by introducing a Dynamic Sparsity MLP, a dedicated neural module that predicts the optimal capacity for every distinct input. Given model activations \(x \in \mathbb{R}^n\), we estimate the input-dependent sparsity level \(\hat{k}\in (0,k_{\max})\) as follows:
\begin{equation}
    \label{k_hat}
    \hat{k} := \sigma(MLP(x)) \cdot k_{\max},
\end{equation}
where \(k_{\max} \in \{1,\ldots, d\}\) is a hyperparameter specifying the upper bound on the sparsity level, $d \in \mathbb{N}^+$ denotes the dictionary size of the autoencoder, and $\sigma$ is the sigmoid activation function. This continuous formulation ensures that the predicted capacity scales smoothly between zero and the prescribed maximum capacity.

We implement the Dynamic Sparsity MLP as a single-hidden-layer network with a hidden dimensionality matching the autoencoder’s latent size \(d\). The first layer is initialized with the encoder weights \(W_{enc}\), thereby providing an inductive bias that aligns the capacity predictor with the feature space used for decomposition, although these parameters are free to diverge during training. We deliberately avoid deeper architectures, as the encoder itself remains linear and therefore cannot exploit higher-order nonlinear representations produced by a deep MLP, making additional depth unnecessary. Despite its simplicity, this design still offers a key advantage over global thresholding approaches: the decision about how many features to activate is made by a dedicated module, disentangled from the reconstruction process, allowing for a more principled and input-dependent allocation of representational capacity.

\begin{figure}
  \centering
  \includegraphics[width=1\textwidth]{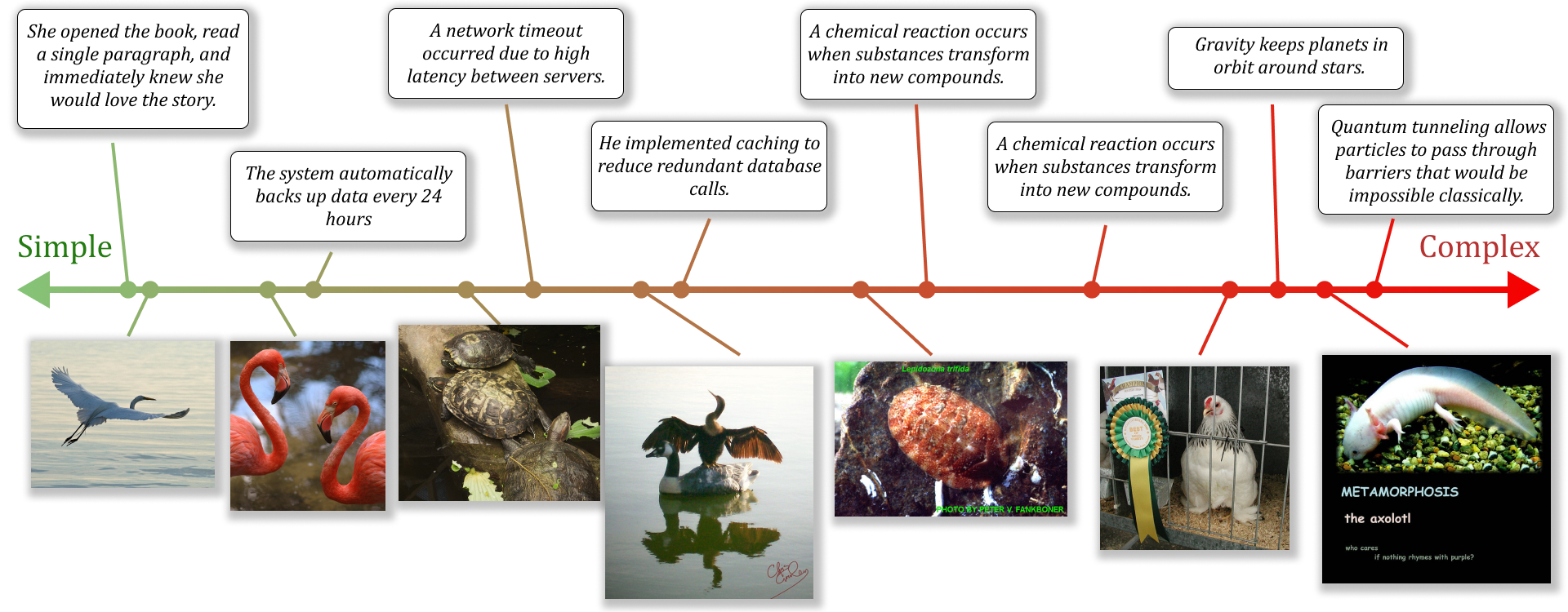}
  \caption{Visualization of inputs with varying embedding complexity: images with a solid background and a single entity lead to a low score, while text and multiple entities produce a high score. For text, the more context and knowledge required to fully capture the meaning of a sentence, the higher the score.}
  \label{complexity_spectrum}
\end{figure}

\paragraph{Differentiable Dynamic Selection}

The central innovation of our methodology lies in integrating the predicted continuous sparsity \(\hat{k}\) directly into the feature selection process. Since standard sorting and Top-K operations are discrete and obstruct gradient flow, we instead employ a differentiable Soft Top-K mechanism during training, while reverting to exact Top-K selection at inference time. Concretely, given latent activations \(z \in \mathbb{R}^d\), the Soft Top-K operator as defined by Struski et al. \cite{struski2025lapsum} (see Section~3.3 therein) produces a vector of continuous selection weights \(p \in (0,1)^d\) such that \(\sum_{i}{p_i} = \hat{k}\). These weights approximate a binary selection mask and are applied elementwise to the latent representation, yielding a soft approximation of Top-K routing.

The sharpness of this approximation is controlled by a temperature parameter \(\alpha>0\) (see \cite{struski2025lapsum}), where values closer to zero recover the hard ranking in the limit. In practice, \(\alpha\) can either be fixed to a small value or gradually annealed toward zero during training; empirically, we observe that such annealing yields modest but consistent improvements. However, in both cases, as $\alpha$ approaches 0, the operator becomes increasingly numerically unstable, producing weights that are nearly binary. While desirable for faithful selection, this regime also introduces a failure mode in which the model can “hide” information in features assigned extremely small but nonzero weights, effectively bypassing the intended sparsity constraint. To mitigate this, we adopt a stabilization strategy in the final phase of training: we freeze the Dynamic Sparsity MLP and replace the soft selection with a hard Top-K operator, ensuring that the model converges to a solution that respects exact, discrete feature selection.
In summary, the forward pass of our model is defined as follows:

{\bf Stage 1.} We first pass the input through the encoder:
\begin{equation}
    z = \text{ReLU}(W_{enc}(x-b_{pre}) + b_{enc}).
\end{equation}

{\bf Stage 2.} During training, we calculate the representation using Soft Top-K, the activations \(z\), the hyperparameter \(\alpha\), and \(\hat{k}\) obtained from \Cref{k_hat}. Notice that we can use a \(\hat{k}\) as a continuous value directly since the Soft Top-K operator naturally accepts non-integer sparsity levels:
\begin{equation}
    p(z) = \text{SoftTopK}(z, \hat{k}, \alpha) \in (0,1)^d,
\end{equation}
\begin{equation}
    f_{train}(z) = p(z) \odot z.
\end{equation}

{\bf Stage 3.} During inference and in the late stages of training, the representation is obtained using the standard hard Top-K operator, with \(\hat{k}\) discretized (via rounding) to meet the operator input requirements.

\begin{equation}
    f_{inf}(z) = \text{TopK}(z, \hat{k}).
\end{equation}

{\bf Stage 4.} Finally, the selected sparse features are projected back into the original space, and the reconstruction follows the standard protocol used in conventional autoencoders:
\begin{equation}
    \hat{x} = W_{dec}f_{inf}(z) + b_{pre}.
\end{equation}


\paragraph{Training Objective and Sparsity Budget}
To effectively train \our{}, we formulate a comprehensive objective function that balances representation quality against our dynamic capacity mechanism. We use a three-term loss consisting of a reconstruction penalty, a sparsity budget constraint, and an optional auxiliary component to maintain neuron health. Consequently, the total loss is defined as:
\begin{equation}
    \mathcal{L}(x) := ||x - \hat{x}(f_{train}(x))||^2_2 + \lambda\mathcal{S}(\hat{k}) + \gamma\mathcal{L}_{aux},
\end{equation}
with all terms detailed in the following paragraphs.

The first term minimizes the standard squared distance between the input data and its reconstruction. The second term, weighted by $\lambda$, regularizes the dynamic sparsity predictor, preventing it from greedily activating the entire dictionary. A natural naive formulation is as follows:

\begin{equation}
\mathcal{S}(\hat{k}) = \text{ReLU}(\mathbb{E}[\hat{k}]-k),
\end{equation}
where \(k < d\) is the hyperparameter that controls the mean sparsity level and \(\mathbb{E}[\cdot]\) denotes the expectation operator. This formulation highlights the core principle behind our sparsity control mechanism: it enforces a budget on the expected number of active features. As long as the model operates below the prescribed threshold, it is free to allocate capacity without incurring any penalty, focusing entirely on minimizing reconstruction error. However, once the expected sparsity \(\mathbb{E}[\hat{k}]\) exceeds this budget, the penalty term becomes active, forcing the model to trade off reconstruction fidelity against the cost of using additional features. In effect, this creates a dynamic equilibrium in which the model must learn to utilize its representational capacity efficiently, activating more neurons only when the corresponding gain in reconstruction quality justifies the incurred penalty.

Despite this appealing interpretation, the use of the ReLU function introduces a significant optimization challenge. Specifically, ReLU is non-differentiable at zero and exhibits a discontinuous gradient, switching abruptly from zero to one. In our setting, this is particularly problematic because the model is explicitly encouraged to operate near the boundary constraint \(\mathbb{E}[\hat{k}] \approx k\), where the penalty transitions from inactive to active. As a result, the optimization landscape becomes sharply kinked in precisely the region most relevant for learning, leading to unstable gradient signals and potentially hindering convergence. This motivates the need for a smoother alternative that preserves the notion of a sparsity budget while providing well-behaved gradients near the constraint boundary. To address these issues, we replace the hard ReLU penalty with a smooth approximation based on the Softplus operator, yielding the final form of our sparsity penalty:

\begin{equation}
    \mathcal{S}(\hat{k}) = \frac{1}{\beta}\ln(1 + e^{(\mathbb{E}[\hat{k}] - k) \cdot \beta}),
\end{equation}
where \(\beta > 0\) controls the sharpness of the transition. If \(\beta \to \infty\), this formulation recovers the original ReLU penalty, while for finite \(\beta\) it provides a continuously differentiable alternative. Crucially, this smoothness eliminates the abrupt change in gradients around the boundary constraint \(\mathbb{E}[\hat{k}] \approx k\), resulting in a more stable optimization landscape. Instead of a hard kink, the model now experiences a gradual increase in penalty as it approaches and exceeds the sparsity budget, allowing gradients to guide it more reliably toward the desired operating point.

\begin{wrapfigure}{r}{0.6\textwidth}
  \centering
  \includegraphics[width=0.6\textwidth]{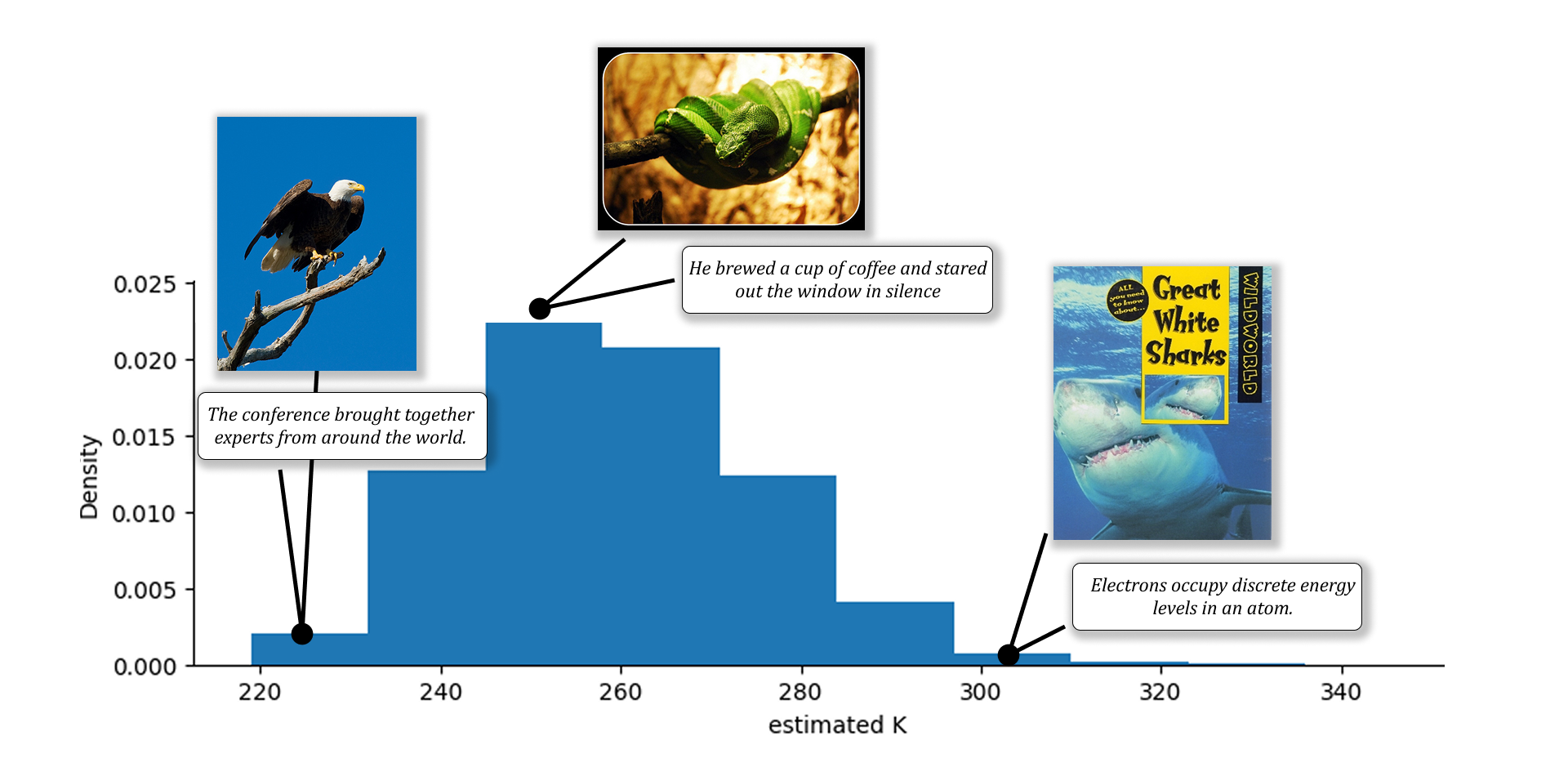}
  \caption{Estimated distribution of \(\hat{k}\), with examples of corresponding texts and images illustrating various levels of complexity.}
  \label{histogram}
\end{wrapfigure}

An additional, somewhat counterintuitive benefit of this formulation emerges in practice. Unlike ReLU, Softplus assigns a small but nonzero penalty even when \(\mathbb{E}[\hat{k}] < k \), effectively discouraging the model from saturating the available capacity. While this may initially appear undesirable, we empirically observe that models trained without such a margin tend to overshoot the intended sparsity level at inference time, activating more features than prescribed. The softened constraint therefore acts as a regularizer that encourages the model to remain conservatively within budget during training, leading to better adherence to the target sparsity level at test time.

The final (optional) component of our objective is an auxiliary loss designed to prevent the emergence of ``dead'' neurons, \marcin{i.e.,} latent features that are rarely or never activated during training. This technique is standard in sparse autoencoder literature and was first introduced in \cite{gao2024scaling}. We follow the conventional formulation:
\begin{equation}
    \mathcal{L}_{aux} = ||(x - \hat{x}) - \hat{e}||^2_2,
\end{equation}
where \(\hat{e}\) denotes a partial reconstruction formed exclusively from the top \(k_{aux}\) latent features that are currently underutilized. Intuitively, this objective encourages these inactive features to capture residual information not already explained by the primary reconstruction. By explicitly assigning them responsibility for modeling the remaining error, the auxiliary loss promotes more balanced utilization of the dictionary and reduces the risk of feature collapse. As a result, the model maintains a healthier and more expressive latent space, in which a larger fraction of neurons contribute meaningfully to the representation.

\section{Experiments}\label{sec:exp}
We evaluate the \our{} model on two distinct modalities to assess both its generality and effectiveness: vision embeddings from CLIP~\cite{clip_matryoshka} and language model activations from Gemma-2-2B \cite{karvonen2025saebench}.
For the vision setting, we extract image embeddings from a pretrained CLIP encoder (ViT-B/16) and train our model directly on these representations, following the setup of \cite{clip_matryoshka}. For the language setting, we operate on post-residual activations from layer 12 of Gemma-2-2B, following standard practices in mechanistic interpretability.

For CLIP SAEs, we use CC3M for training, and Imagenet-1k/Imagenet-100 \cite{deng2009imagenet} for evaluation. The dictionary size is set to \(d=4096\). All CLIP evaluations are conducted across five different target sparsity levels, \(k \in \{60, 100, 140, 180, 220\} \).

Gemma-2-2B SAEs are trained on the FineWeb dataset \cite{penedo2024the}, with the dictionary size set to \(d=2^{14}=16384\). All Gemma-2-2B evaluations are performed across five different target sparsity levels, \(k \in \{20, 40, 80, 160, 320\}\).

Across all experiments, we compare \our{} against several well-established architectures: TopK SAE \cite{gao2024scaling}, BatchTopK SAE \cite{bussmann2024batchtopk}, and the recent Matryoshka SAE \cite{clip_matryoshka, bussmann2025learning}. Evaluation focuses on three axes: reconstruction fidelity, sparsity control, and usefulness of the learned features.

All training and experiments were conducted on a single Nvidia GH200 120GB GPU with 64GB RAM within a SLURM cluster. For CLIP, we trained for 40k steps, requiring approximately 2 hours for \our{} and 1.5 hours for the baselines. For Gemma-2-2B, we trained for approximately 150k steps, requiring around 6 hours for \our{} and 2.5 hours for the baselines. Hyperparameter settings were selected through an extensive Bayesian sweep and are reported in detail in Appendix \ref{app:d}.

\begin{figure}
  \centering
  \includegraphics[width=0.45\textwidth]{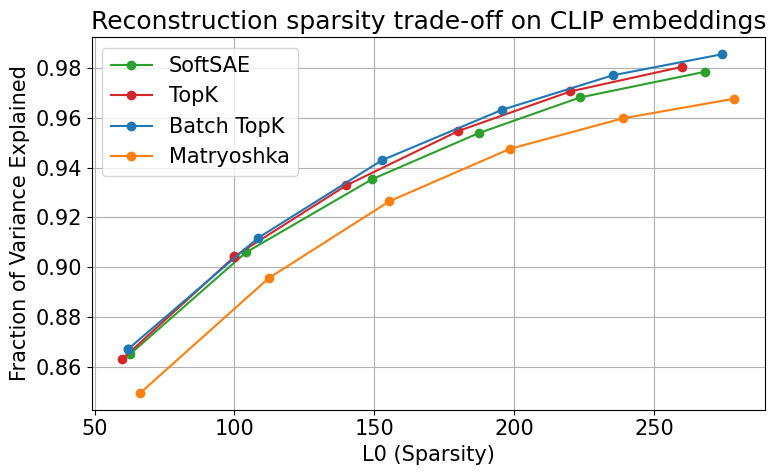}
  \includegraphics[width=0.48\textwidth]{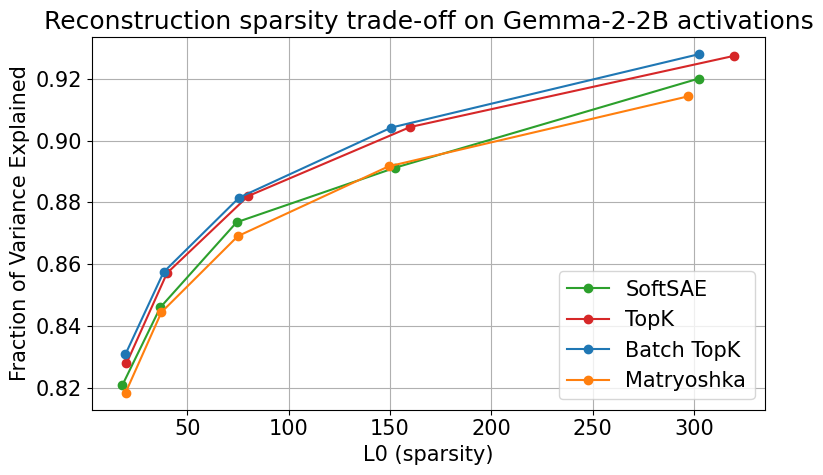}
  \caption{Reconstruction quality versus sparsity level for different model architectures evaluated on ImageNet CLIP embeddings (left) and Gemma-2-2B 12th layer activations (right). \our{} achieves tight sparsity control with competitive reconstruction fidelity.}
  \label{rec_sparsity}
\end{figure}

\paragraph{Reconstruction--Sparsity Trade-off}
A key objective of sparse autoencoders is to balance reconstruction fidelity with controllable sparsity, which we evaluate using the fraction of explained variance (\Cref{rec_sparsity}). On CLIP embeddings, \our{} matches the reconstruction--sparsity performance of TopK and BatchTopK, slightly outperforming BatchTopK at comparable error while remaining close to TopK, which retains a small edge. This proximity is important because TopK methods, while effective at optimizing the trade-off, often do so at the expense of feature quality; \our{} instead sacrifices only a negligible amount of reconstruction performance in exchange for improved feature quality, similar to Matryoshka representations. On Gemma-2-2B activations, \our{} performs comparably to Matryoshka, with a slight advantage in reconstruction at similar sparsity levels. Overall, \our{} remains competitive on the reconstruction–sparsity frontier while offering clear gains in feature structure and interpretability.

\paragraph{Feature Quality Evaluation (SAE Bench)}
Beyond reconstruction, we assess interpretability and utility using selected modules from SAE Bench \cite{karvonen2025saebench}, focusing on feature splitting, feature absorption, spurious correlation removal, and targeted probe perturbation. These diagnostics provide a finer-grained view of representation quality by directly probing the structure, robustness, and disentanglement of learned features.

\begin{figure}
  \centering
  \includegraphics[width=0.45\textwidth]{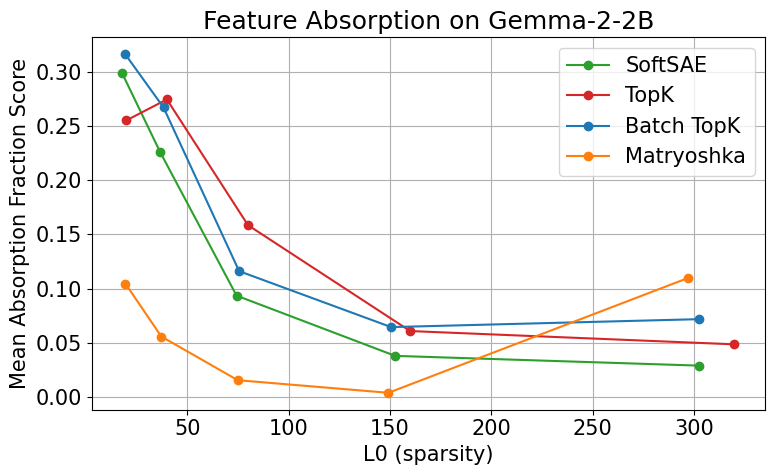}
  \includegraphics[width=0.45\textwidth]{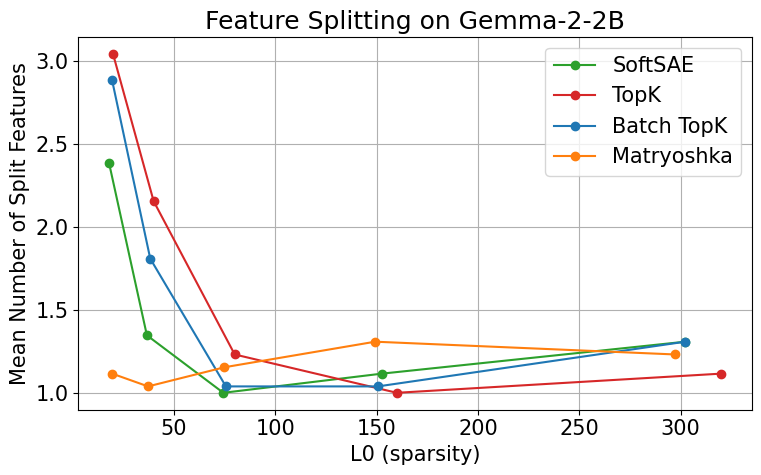}
  \caption{Absorption Rate and Number of Split Features vs. Sparsity. \our{} demonstrates its resistance to feature absorption and splitting anomalies.}
  \label{absorption_splitting}
\end{figure}

Using SAE Bench absorption and splitting metrics (\Cref{absorption_splitting}), which measure whether semantic features are overly concentrated (absorption) or redundantly distributed (splitting), we observe that \our{} performs comparably to state-of-the-art Matryoshka SAEs while showing a clear advantage in reducing feature absorption at higher \(L_0\) sparsity metric value. This indicates better preservation of general semantic features as representations become more specialized.

\begin{figure}
  \centering
  \includegraphics[width=0.45\textwidth]{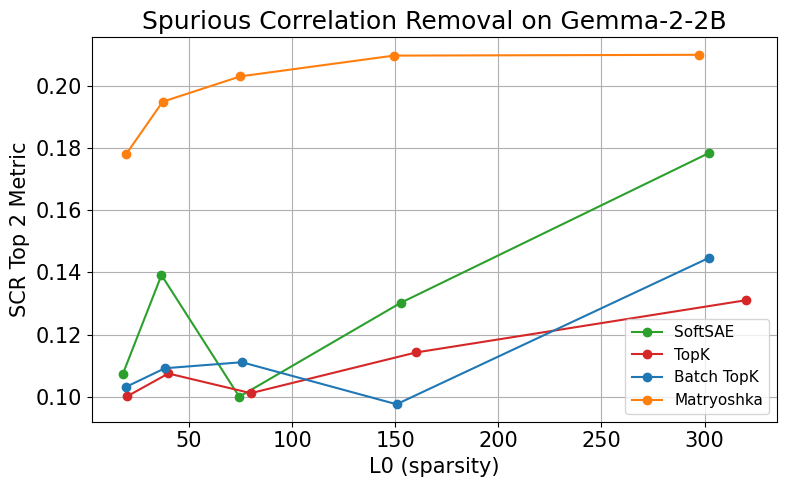}
  \includegraphics[width=0.45\textwidth]{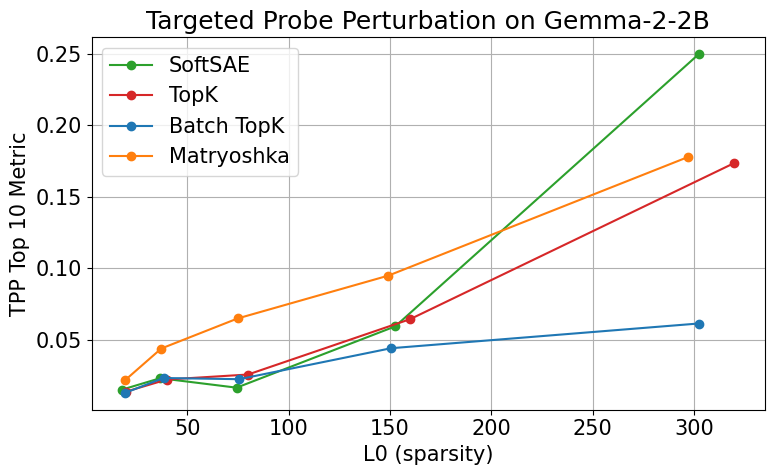}
  \caption{Spurious Correlation Removal and Targeted Probe Perturbation scores vs. Sparsity. \our{} proves effective at feature disentanglement, especially at higher \(L_0\) values.}
  \label{scr_tpp}
\end{figure}

\paragraph{Sparse Probing and Targeted Concept Removal}
We further evaluate representation quality with Spurious Correlation Removal (SCR) and Targeted Probe Perturbation (TPP) (\Cref{scr_tpp}), which assess semantic alignment, robustness to spurious features, and disentanglement. \our{} outperforms TopK and BatchTopK on SCR and remains competitive with Matryoshka, while matching and at higher sparsity level surpassing Matryoshka on TPP, indicating strong disentanglement and precise, localized control over learned features.


\paragraph{Qualitative Analysis of Dynamic Sparsity}

    

We visualize the behavior of the Dynamic Sparsity MLP in \Cref{complexity_spectrum,histogram}, with additional examples and details provided in Appendix \ref{app:a}. We observe that inputs with low complexity, such as images with a uniform background and a single entity, consistently receive low scores, as they require only a small number of features to represent their content. In contrast, inputs with high complexity, including scenes with multiple entities, rich backgrounds, or textual content, receive higher scores; for text in particular, the estimated complexity increases with the amount of context and knowledge required to fully capture the meaning.

\section{Conclusions}
In this work, we introduced \our{}, a sparse autoencoder architecture that replaces fixed-capacity constraints with a fully differentiable, input-adaptive sparsity mechanism. While maintaining competitive performance on the reconstruction--sparsity trade-off relative to strong baselines, SoftSAE consistently yields well structured and disentangled features, particularly at higher sparsity levels. Importantly, its primary contribution is not merely improved reconstruction or feature quality, but the ability to dynamically allocate representational capacity based on input complexity. This dynamic sparsity control is learned end-to-end and correlates closely with both visual and textual complexity, enabling the model to produce shorter, cleaner explanations for simple inputs and richer representations for complex ones. Despite current limitations, most notably increased computational cost and the need to carefully manage the soft-to-hard selection transition, our approach opens a new direction in which sparsity is no longer a fixed hyperparameter but a flexible, learnable component. Because the mechanism is fully differentiable, it can be seamlessly integrated into broader neural systems, including generative pipelines, where adaptive allocation of capacity may improve both efficiency and interpretability.

\paragraph{Limitations}
\our{} incurs higher computational cost than baseline methods, primarily due to the differentiable Soft Top-K operator, which becomes a bottleneck for large dictionaries (e.g., with \(d=2^{16}\)) and may limit scalability. 

\medskip
\bibliographystyle{unsrt}

\clearpage
\appendix

\section{Impact Statement and Declaration of LLM Usage}

Our work contributes to the safety and alignment of Large Language Models (LLMs) and Vision Transformers (ViTs) by providing more faithful and adaptive internal representations. While we acknowledge that interpretability tools could theoretically be used to exploit model vulnerabilities, we emphasize that SoftSAE primarily serves as a foundational tool for defensive safety research.

LLMs were utilized exclusively for the purposes of language editing and text refinement to improve the clarity of the manuscript.

\section{Qualitative Examples of Extreme Sparsity Regimes}\label{app:a}

\begin{figure}[H]
  \centering
  \includegraphics[width=1\textwidth]{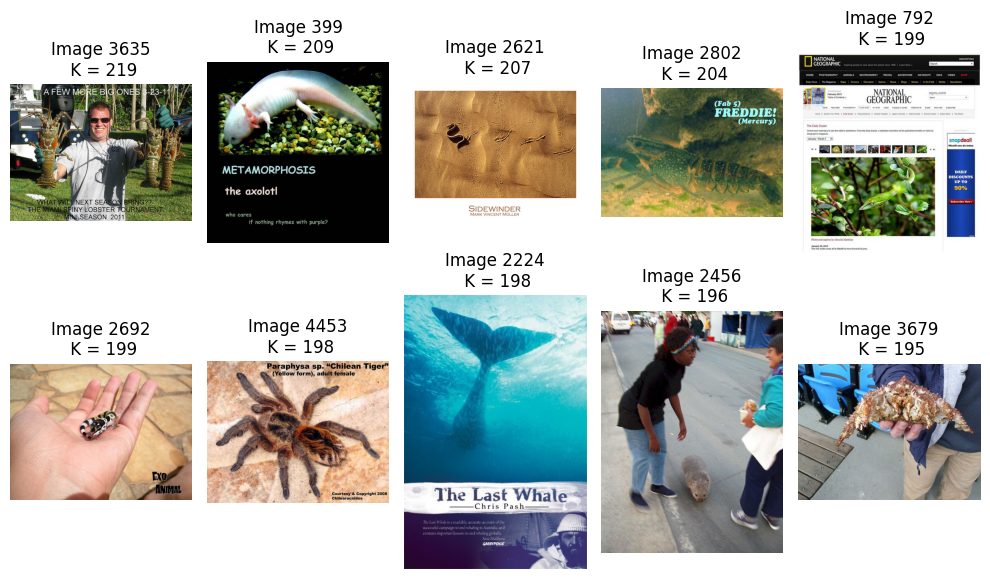}
  \caption{Examples of images from ImageNet100 with the highest estimated \(\hat{k}\) values (target \(K=140\)); most of them include text or at least multiple objects.}
  \label{clip_high_complexity}
\end{figure}

\begin{figure}[]
  \centering
  \includegraphics[width=1\textwidth]{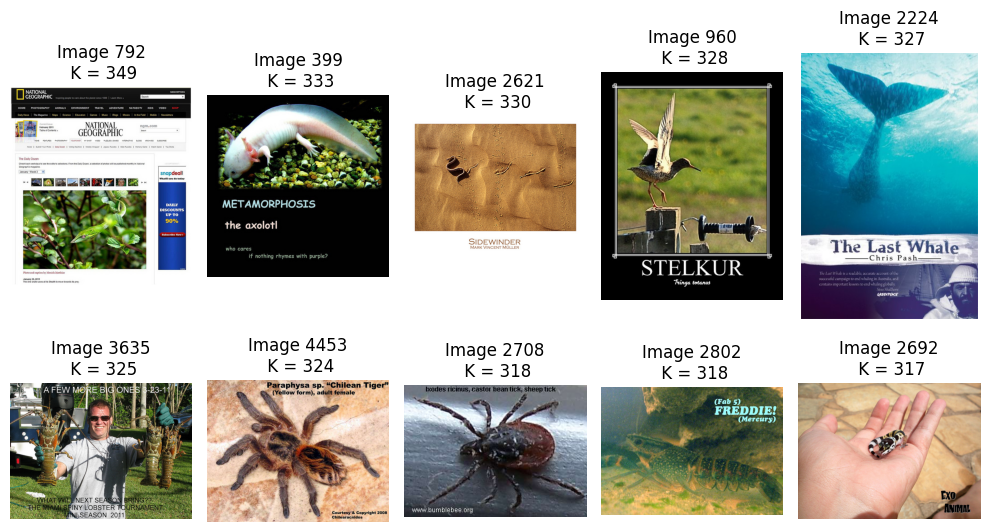}
  \caption{Additional examples of images from ImageNet100 with the highest estimated \(\hat{k}\) values (target \(K=260\)); most of them include text or at least multiple objects.}
  \label{clip_high_complexity_260}
\end{figure}

\begin{figure}[]
  \centering
  \includegraphics[width=1\textwidth]{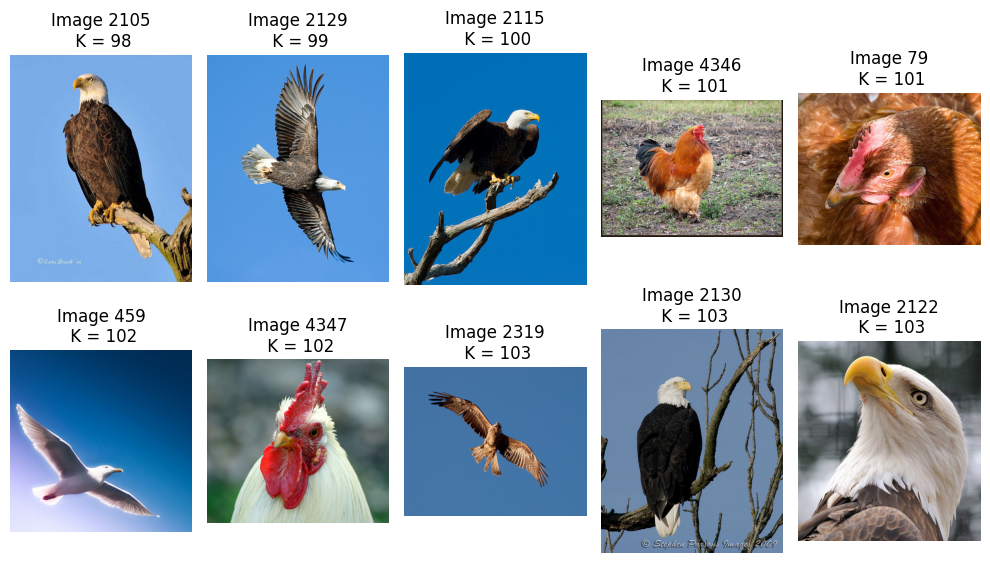}
  \caption{Examples of images from ImageNet100 with the lowest estimated \(\hat{k}\) values (target \(K=140\)); most of them contain a single entity and have no background or a plain background.}
  \label{clip_low_complexity}
\end{figure}

\begin{figure}[]
  \centering
  \includegraphics[width=1\textwidth]{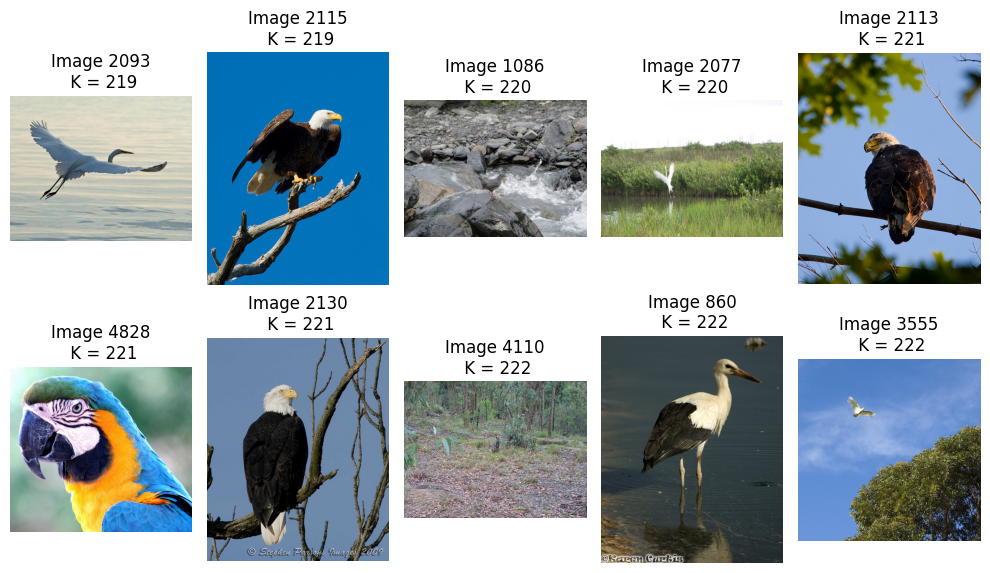}
  \caption{Additional examples of images from ImageNet100 with the lowest estimated \(\hat{k}\) values (target \(K=260\)); most of them contain a single entity and have no background or a plain background.}
  \label{clip_low_complexity_260}
\end{figure}

To better illustrate the behavior of the Dynamic Sparsity mechanism, we present additional qualitative examples corresponding to extreme values of the estimated sparsity level \(\hat{k}\) (see \Cref{clip_high_complexity,clip_high_complexity_260,clip_low_complexity,clip_low_complexity_260}). These examples highlight how the model adapts its representational capacity to the intrinsic complexity of the input.

At the low end of the spectrum, we observe inputs assigned very small \(\hat{k}\) values. These are typically simple images containing a single, well-isolated object on a clean or uniform background. In such cases, only a few features are sufficient to reconstruct the representation, leading to short and highly focused explanations. Notably, many examples in this regime include images of birds, suggesting that the concept of a bird in CLIP embeddings is highly separable and can be captured with a small number of distinct features.

At the high end of the spectrum, inputs with large \(\hat{k}\) values exhibit significantly greater complexity. These include scenes with multiple objects, cluttered backgrounds, or the presence of text. In particular, images containing text consistently receive higher scores, indicating that textual content introduces additional semantic dimensions that require more features to represent accurately. Similarly, multi-object compositions demand a richer set of features to disentangle overlapping concepts and interactions.

Overall, these examples reinforce a key property of our approach: the estimated sparsity level \(\hat{k}\) aligns closely with intuitive notions of input complexity. Simple, single-object scenes yield compact representations, while visually or semantically dense inputs naturally expand the number of active features.

\section{Concept Masking}
\begin{figure}[]
  \centering
  \includegraphics[width=0.8\textwidth]{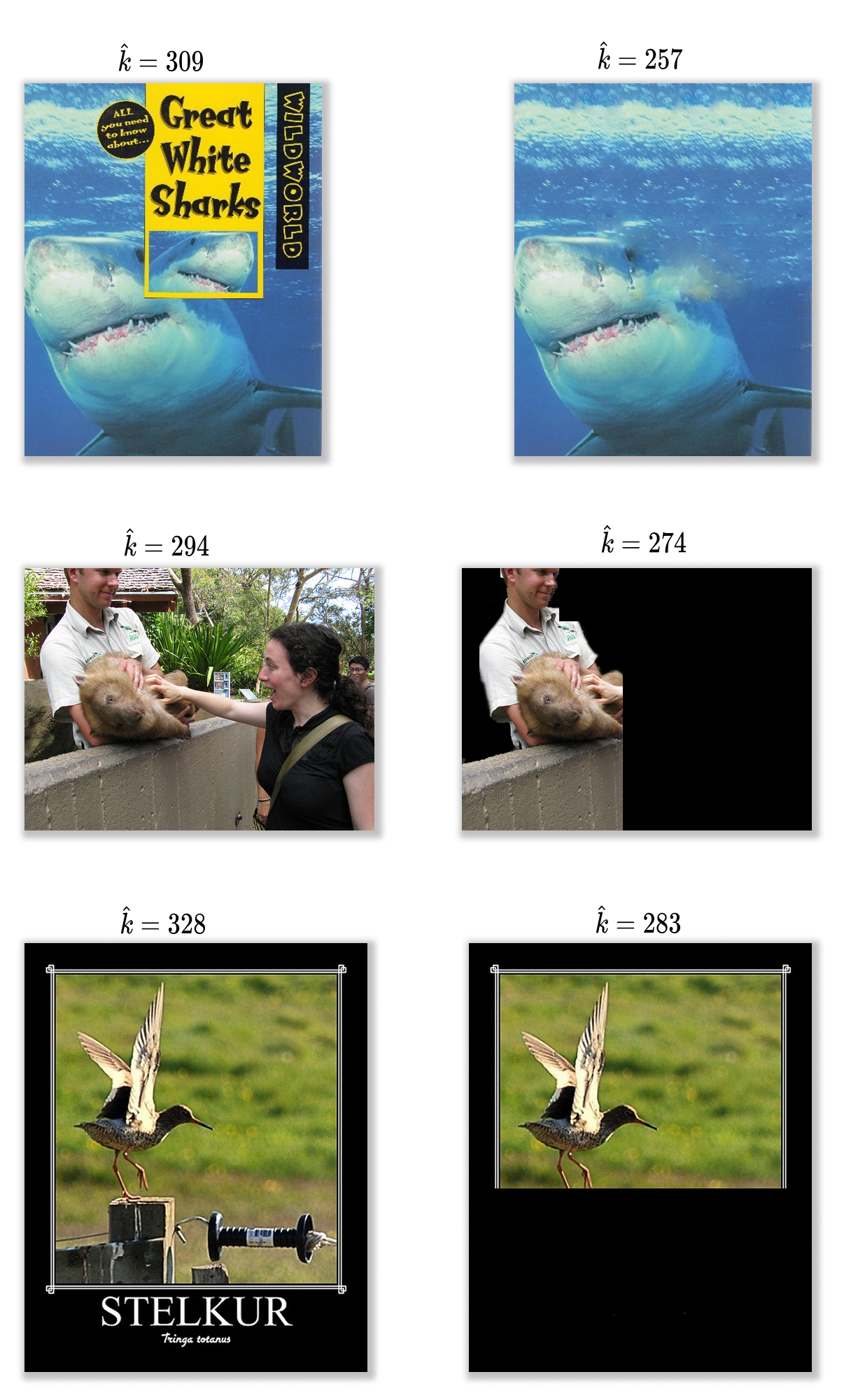}
  \caption{The presented results indicate that the Dynamic Sparsity MLP is sensitive not only to the presence of information, but also to the manner in which it is removed. Naive masking with solid black rectangles consistently reduces the predicted sparsity \(\hat{k}\), confirming that the model reacts to a loss of visual content. However, this approach proves less effective than smoother, structure-preserving transformations. In particular, methods that remove or simplify content without introducing artificial high-contrast artifacts, such as background-aware inpainting, gradual blurring, or cropping that respects existing image boundaries, lead to a more pronounced decrease in \(\hat{k}\). For example, when images contain natural dark borders, simply expanding these regions inward reduces the complexity score more reliably than inserting abrupt occlusions into high-frequency areas. This suggests that the model partially interprets sharp artificial edges as informative signals, whereas more naturalistic simplifications better align with its notion of reduced semantic complexity.}
  \label{masking}
\end{figure}

To gain deeper insight into how the Dynamic Sparsity MLP estimates input complexity, we conduct a controlled perturbation analysis on high-complexity images. Starting from samples that yield high predicted sparsity levels (\(\hat{k}\)), we systematically simplify their visual content by removing or degrading informative components. Results are shown in \Cref{masking}.

\section{Detailed Numerical Results}
To complement the visual comparisons presented in the figures, we additionally report detailed numerical results in \Cref{clip_table,gemma_table}. While plots provide an intuitive overview of trends and trade-offs, they can obscure precise quantitative differences between methods, particularly at specific sparsity levels. The tables, therefore, present exact values for reconstruction quality, sparsity metrics, and interpretability benchmarks across all evaluated configurations. This allows for a more fine-grained comparison between \our{} and baseline approaches, highlighting subtle performance gaps and confirming consistency of the observed trends.

\begin{table}[]
\caption{Detailed numerical results from experiments on CLIP embeddings.}
\label{clip_table}
\centering
\begin{tabular}{@{}llll@{}}
\toprule
\textbf{Architecture} & \textbf{k} & \textbf{L0}      & \textbf{FVE}   \\ \midrule
TopK                  & 60         & \textbf{60.000}  & 0.863          \\
Batch TopK            & 60         & {\ul 61.906}     & \textbf{0.867} \\
Matryoshka            & 60         & 66.420           & 0.850          \\
SoftSAE               & 60         & 62.562           & {\ul 0.865}    \\
                      &            &                  &                \\
TopK                  & 100        & \textbf{100.000} & 0.904          \\
BatchTopK             & 100        & 108.314          & \textbf{0.912} \\
Matryoshka            & 100        & 112.275          & 0.896          \\
SoftSAE               & 100        & {\ul 104.150}    & {\ul 0.906}    \\
                      &            &                  &                \\
TopK                  & 140        & \textbf{140.000} & 0.933          \\
Batch TopK            & 140        & 152.770          & \textbf{0.943} \\
Matryoshka            & 140        & 155.405          & 0.926          \\
SoftSAE               & 140        & {\ul 149.300}    & {\ul 0.935}    \\
                      &            &                  &                \\
TopK                  & 180        & \textbf{180.000} & {\ul 0.955}    \\
Batch TopK            & 180        & 195.747          & \textbf{0.963} \\
Matryoshka            & 180        & 198.477          & 0.947          \\
SoftSAE               & 180        & {\ul 187.240}    & 0.954          \\
                      &            &                  &                \\
TopK                  & 220        & \textbf{220.000} & {\ul 0.970}    \\
Batch TopK            & 220        & 235.186          & \textbf{0.977} \\
Matryoshka            & 220        & 238.797          & 0.960          \\
SoftSAE               & 220        & {\ul 223.459}    & 0.968          \\
                      &            &                  &                \\
TopK                  & 260        & \textbf{260.000} & {\ul 0.980}    \\
Batch TopK            & 260        & 274.274          & \textbf{0.985} \\
Matryoshka            & 260        & 278.693          & 0.968          \\
SoftSAE               & 260        & {\ul 268.338}    & 0.978\\      \bottomrule   
\end{tabular}
\end{table}

\begin{table}[]
\caption{Detailed numerical results from experiments on Gemma-2-2B activations.}
\label{gemma_table}
\centering
\begin{tabular}{@{}llrrrrrr@{}}
\toprule
\textbf{Architecture} & \textbf{k} & \multicolumn{1}{l}{\textbf{L0}} & \multicolumn{1}{l}{\textbf{FVE}} & \multicolumn{1}{l}{\textbf{Absorp Frac}} & \multicolumn{1}{l}{\textbf{Splitting}} & \multicolumn{1}{l}{\textbf{SCR TOP 2}} & \multicolumn{1}{l}{\textbf{TPP TOP 10}} \\ \midrule
Batch TopK            & 20         & {\ul 19.452}                    & \textbf{0.831}                   & 0.317                                            & 2.885                                  & 0.103                                  & 0.013                                   \\
Matryoshka            & 20         & 19.531                          & 0.818                            & \textbf{0.105}                                   & \textbf{1.115}                         & \textbf{0.178}                         & \textbf{0.022}                          \\
TopK                  & 20         & 20.000                          & {\ul 0.828}                      & {\ul 0.255}                                      & 3.038                                  & 0.100                                  & 0.014                                   \\
SoftSAE               & 20         & \textbf{17.977}                 & 0.821                            & 0.299                                            & {\ul 2.385}                            & {\ul 0.107}                            & {\ul 0.015}                             \\
                      &            &                                 &                                  &                                                  &                                        &                                        &                                         \\
Batch TopK            & 40         & 38.326                          & \textbf{0.857}                   & 0.267                                            & 1.808                                  & 0.109                                  & {\ul 0.023}                             \\
Matryoshka            & 40         & {\ul 37.247}                    & 0.845                            & \textbf{0.056}                                   & \textbf{1.038}                         & \textbf{0.195}                         & \textbf{0.044}                          \\
TopK                  & 40         & 40.000                            & \textbf{0.857}                   & 0.275                                            & 2.154                                  & 0.108                                  & 0.022                                   \\
SoftSAE               & 40         & \textbf{36.668}                 & {\ul 0.846}                      & {\ul 0.226}                                      & {\ul 1.346}                            & {\ul 0.139}                            & {\ul 0.023}                             \\
                      &            &                                 &                                  &                                                  &                                        &                                        &                                         \\
Batch TopK            & 80         & 75.649                          & \textbf{0.882}                   & 0.116                                            & {\ul 1.038}                            & {\ul 0.111}                            & 0.023                                   \\
Matryoshka            & 80         & {\ul 74.956}                    & 0.869                            & \textbf{0.015}                                   & 1.154                                  & \textbf{0.203}                         & \textbf{0.065}                          \\
TopK                  & 80         & 80.000                            & \textbf{0.882}                   & 0.159                                            & 1.231                                  & 0.101                                  & {\ul 0.026}                             \\
SoftSAE               & 80         & \textbf{74.380}                  & {\ul 0.874}                      & {\ul 0.093}                                      & \textbf{1.000}                         & 0.100                                  & 0.017                                   \\
                      &            &                                 &                                  &                                                  &                                        &                                        &                                         \\
Batch TopK            & 160        & {\ul 150.634}                   & \textbf{0.904}                   & 0.064                                            & {\ul 1.038}                            & 0.098                                  & 0.044                                   \\
Matryoshka            & 160        & \textbf{149.306}                & 0.892                            & \textbf{0.004}                                   & 1.308                                  & \textbf{0.210}                         & \textbf{0.095}                          \\
TopK                  & 160        & 160.000                           & \textbf{0.904}                   & 0.060                                            & \textbf{1.000}                         & 0.114                                  & {\ul 0.065}                             \\
SoftSAE               & 160        & 152.657                         & 0.891                            & {\ul 0.038}                                      & 1.115                                  & {\ul 0.130}                            & 0.060                                    \\
                      &            &                                 &                                  &                                                  &                                        &                                        &                                         \\
Batch TopK            & 320        & {\ul 302.343}                   & \textbf{0.928}                   & 0.072                                            & 1.308                                  & 0.145                                  & 0.062                                   \\
Matryoshka            & 320        & \textbf{297.323}                & 0.914                            & 0.110                                            & {\ul 1.231}                            & \textbf{0.210}                         & {\ul 0.178}                             \\
TopK                  & 320        & 320.000                           & {\ul 0.927}                      & {\ul 0.048}                                      & \textbf{1.115}                         & 0.131                                  & 0.174                                   \\
SoftSAE               & 320        & 302.355                         & 0.920                             & \textbf{0.029}                                   & 1.308                                  & {\ul 0.178}                            & \textbf{0.250} \\
\bottomrule
\end{tabular}
\end{table}

\section{Hyperparameter Settings}\label{app:d}
We provide detailed hyperparameter configurations used across all experiments for reproducibility. \Cref{clip_hyper,gemma_hyper} include settings for both vision (CLIP) and language (Gemma) models, covering optimization, sparsity control, and architectural choices. While most parameters remain consistent across runs, key values such as target sparsity \(K\), learning rates, and regularization strengths were tuned to ensure a fair comparison with baseline methods.

\begin{table}[]
\centering
\small

\begin{minipage}[t]{0.48\textwidth}
\centering
\caption{Detailed hyperparameter settings in CLIP experiments.}
\label{clip_hyper}
\begin{tabular}{lr}
\toprule
\textbf{Category} & \textbf{Setting} \\
\midrule

\multicolumn{2}{l}{\textit{Model / Architecture}} \\
Activation dimension & 512 \\
Dictionary size      & 4096 \\
Model name           & CLIP \\

\midrule
\multicolumn{2}{l}{\textit{Optimization}} \\
Learning rate        & 6e-4 \\
Total steps          & 40{,}000 \\
Warmup steps         & 1{,}900 \\
Decay start          & 6{,}500 \\

\midrule
\multicolumn{2}{l}{\textit{Top-K Parameters}} \\
$k_{\text{max}}$     & \(2 \cdot k\) \\
Auxiliary $k$        & 256 \\
Hard top-$k$ steps   & 6{,}000 \\

\midrule
\multicolumn{2}{l}{\textit{Loss \& Regularization}} \\
$k$ loss weight      & 1.0 \\
$k$ loss beta        & 5.0 \\
Auxiliary $\alpha$   & 0.1 \\
Soft top-$k$ $\alpha$ & 1e-4 \\

\midrule
\multicolumn{2}{l}{\textit{Annealing}} \\
$k$ anneal steps     & 1{,}600 \\
$\alpha$ anneal steps & 1{,}000 \\

\midrule
\multicolumn{2}{l}{\textit{Other}} \\
Dead feature threshold & 400{,}000 \\
\bottomrule
\end{tabular}
\end{minipage}
\hfill
\begin{minipage}[t]{0.48\textwidth}
\centering
\caption{Detailed hyperparameter settings in Gemma-2-2B experiments.}
\label{gemma_hyper}
\begin{tabular}{lr}
\toprule
\textbf{Category} & \textbf{Setting} \\
\midrule

\multicolumn{2}{l}{\textit{Model / Architecture}} \\
Activation dimension & 2304 \\
Dictionary size      & 16384 \\
Model name           & Gemma-2-2B \\

\midrule
\multicolumn{2}{l}{\textit{Optimization}} \\
Learning rate        & 3e-4 \\
Total steps          & 146{,}484 \\
Warmup steps         & 1{,}000 \\
Decay start          & 117{,}187 \\

\midrule
\multicolumn{2}{l}{\textit{Top-K Parameters}} \\
$k_{\text{max}}$     & \(2 \cdot k\) \\
Auxiliary $k$        & 1152 \\
Hard top-$k$ steps   & 10{,}000 \\

\midrule
\multicolumn{2}{l}{\textit{Loss \& Regularization}} \\
$k$ loss weight      & 1.0 \\
$k$ loss beta        & 5.0 \\
Auxiliary $\alpha$   & 0.03125 \\
Soft top-$k$ $\alpha$ & 1e-4 \\

\midrule
\multicolumn{2}{l}{\textit{Annealing}} \\
$k$ anneal steps     & 1{,}464 \\
$\alpha$ anneal steps & 2{,}500 \\

\midrule
\multicolumn{2}{l}{\textit{Other}} \\
Dead feature threshold & 10{,}000{,}000 \\
\bottomrule
\end{tabular}
\end{minipage}

\end{table}

\section{Annealing Ablation}
\begin{figure}[]
  \centering
  \includegraphics[width=0.7\textwidth]{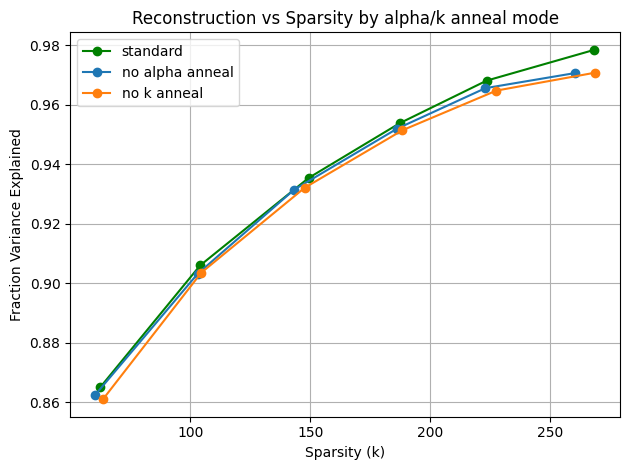}
  \caption{Ablation study within CLIP reconstruction experimental setup. Removing \(\alpha\)/\(k\) annealing results in a minor drop in reconstruction quality.}
  \label{alpha_k_ablation}
\end{figure}

We further investigate the impact of annealing strategies through an ablation study on both the Soft Top-K temperature parameter \(\alpha\) and the target sparsity \(K\) (\Cref{alpha_k_ablation}). Although the improvements are modest, they are consistent across settings, suggesting that annealing both the selection sharpness and the target sparsity provides a better alignment between the soft training dynamics and the hard Top-K behavior used at inference time.

\end{document}